\documentclass[a4paper,fleqn]{cas-dc}

\usepackage[numbers]{natbib}
\usepackage{placeins} % 在导言区加载宏包
\usepackage{caption}
\usepackage{graphicx}
\usepackage{subcaption}

\def\tsc#1{\csdef{#1}{\textsc{\lowercase{#1}}\xspace}}
\tsc{WGM}
\tsc{QE}
\tsc{EP}
\tsc{PMS}
\tsc{BEC}
\tsc{DE}
\begin{document}
\let\WriteBookmarks\relax
\def\floatpagepagefraction{1}
\def\textpagefraction{.001}
\shorttitle{Expert Systems with Applications}
\shortauthors{Guangrui Bai et~al.}

\title [mode = title]{Towards Lightest Low-Light Image Enhancement Architecture for Mobile Devices} 

\author[addr1]{Guangrui~Bai}\fnref{co-first}
\author[addr2]{Hailong~Yan}\fnref{co-first}

\author[addr1]{Wenhai~Liu}
%\ead{wenhailiu@mail.ustc.edu.cn}
\author[addr1]{Yahui~Deng}
%\ead{yanhailong@std.uestc.edu.cn}
\author[addr1]{Erbao~Dong\corref{cor1}}[orcid=0000-0002-4062-9730]
\ead{ebdong@ustc.edu.cn}

\address[addr1]{Key Laboratory of Precision and Intelligent Chemistry, Department of Precision Machinery and Precision Instrumentation, University of Science and Technology of China, Hefei, Anhui 230026, China.}
\address[addr2]{School of Information and Communication Engineering, University of Electronic Science and Technology of China, Chengdu 611731, China.}

\cortext[cor1]{Corresponding author.}
\tnotetext[1]{This work was supported by the National Key R\&D Program of China (Grant No. 2018YFB1307400) and the State Grid Anhui Science and Technology Project.}
\fntext[co-first]{Equal Contribution.}

%\tnotetext[1]{This work was supported by the National Key R\&D Program of China (Grant No. 2018YFB1307400) and the State Grid Anhui Science and Technology Project.}
%\fntext[1]{These authors contributed equally to this work.}

\begin{abstract}
Real-time low-light image enhancement on mobile and embedded devices requires models that balance visual quality and computational efficiency. Existing deep learning methods often rely on large networks and labeled datasets, limiting their deployment on resource-constrained platforms. In this paper, we propose LiteIE, an ultra-lightweight unsupervised enhancement framework that eliminates dependence on large-scale supervision and generalizes well across diverse conditions. We design a backbone-agnostic feature extractor with only two convolutional layers to produce compact image features enhancement  tensors. In addition, we develop a parameter-free Iterative Restoration Module, which reuses the extracted features to progressively recover fine details lost in earlier enhancement steps, without introducing any additional learnable parameters. We further propose an unsupervised training objective that integrates exposure control, edge-aware smoothness, and multi-scale color consistency losses. Experiments on the LOL dataset, LiteIE achieves 19.04 dB PSNR, surpassing SOTA by 1.4 dB while using only 0.07\% of its parameters. On a Snapdragon 8 Gen 3 mobile processor, LiteIE runs at 30 FPS for 4K images with just 58 parameters, enabling real-time deployment on edge devices. These results establish LiteIE as an efficient and practical solution for low-light enhancement on resource-limited platforms. The code is available at: \url{https://github.com/mubaisam/LiteIE}.

\end{abstract}

%\includegraphics[width=0.85\textwidth]{figure1.pdf}
%%\includegraphics[width=0.25\textwidth]{sec/Image/figure2a.pdf} & % 设置宽度为0.48\textwidth，占据左右两栏
%    \item We design a convolution-independent feature extraction framework.
%    \item we develop the lightest low-light enhancement model with 58 parameters.
%    \item A multi-stage aggregation strategy enhances details and contrast with low complexity.
%    \item LiteIE achieves real-time 4K 30FPS on Snapdragon 8 Gen 3 mobile processors.
%    \end{highlights}


\maketitle

\section{Introduction}

\begin{figure*}[t] % 使用 figure* 让图像跨越双栏
    \centering
    %\captionsetup{type=figure}
    \setlength{\tabcolsep}{2pt} % 设置表格列间的间隔
    \begin{tabular}{@{}cc@{}} % 去掉左右边界，显示两幅图
        \includegraphics[width=0.3\textwidth]{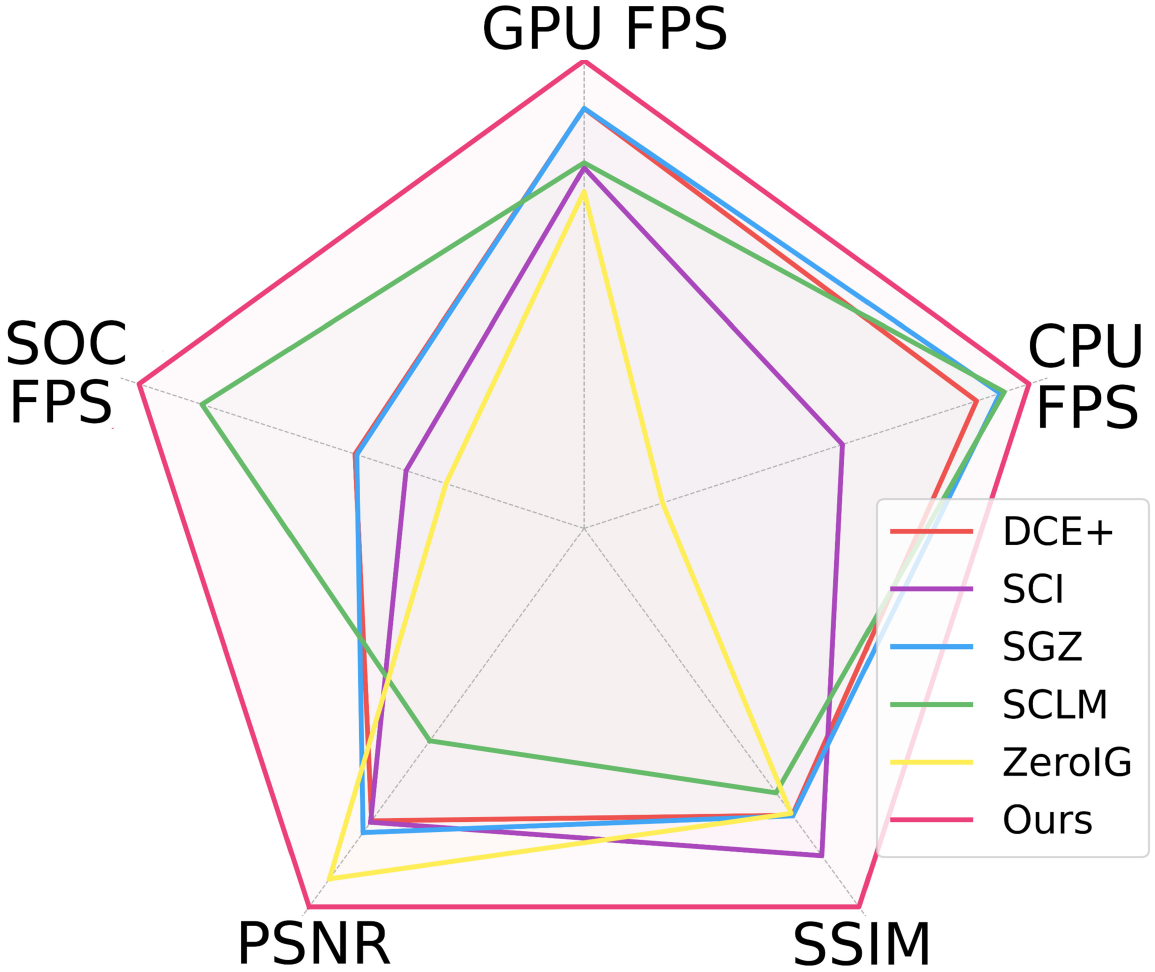} & % 设置宽度为0.48\textwidth，占据左右两栏
        \includegraphics[width=0.7\textwidth]{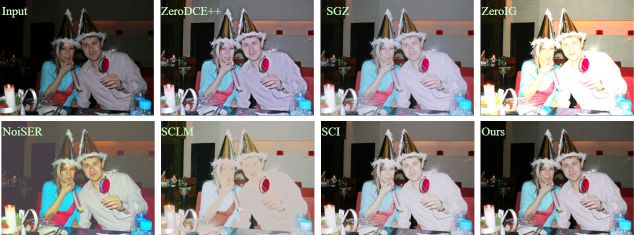} \\ % 调整两图的宽度，避免间隔过大
        (a) Efficiency Analysis & (b) Visual Comparison \\
    \end{tabular}
    \caption{Performance and efficiency comparisons with state-of-the-art methods. (a) Image quality metrics (PSNR, SSIM) and runtime efficiency (CPU, GPU, and mobile phone SoC FPS). Our method consistently outperforms other approaches. (b) Visual results show LiteIE produces natural and perceptually pleasing enhancements.
}
    
\end{figure*}

low-light image enhancement (LLIE) is a key challenge in computer vision \cite{li2021low}, particularly in resource-constrained settings such as autonomous driving, surveillance, and mobile photography. Images captured under low-light conditions often suffer from poor visibility, low contrast, heavy noise, and color distortion \cite{li2021low}, degrading both human perception and downstream tasks (e.g., detection, recognition, tracking). In safety-critical scenarios such as nighttime driving, these issues can lead to perception failures with serious risks. Therefore, designing efficient LLIE methods that balance performance and speed remains a pressing need. 

Existing solutions for low-light imaging can be divided into hardware- and software-based approaches. Hardware methods (e.g., large-aperture lenses, high-sensitivity sensors, multi-exposure fusion) can effectively boost illumination but often incur high cost, increased power consumption, or bulky designs. In contrast, software-based techniques are more flexible and cost-efficient. Traditional histogram equalization \cite{pizer1987adaptive} stretches pixel intensities to improve contrast but easily amplifies noise or causes color shifts. Retinex-based methods \cite{1971retinex, guo2016lime} decompose images into illumination and reflectance to enhance dark regions, but their limited adaptive learning often hinders generalization in complex scenes.

Deep learning has made substantial strides in LLIE, comprising both supervised and unsupervised paradigms \cite{zhou2025cwfas,jiang2025low,yan2025mobileie,liu2024ntire,yanigdnet,yan2025zero}. Supervised methods learn mappings from paired low-/normal-light images~\cite{wu2022uretinex,zhang2019kindling} but often overfit to specific training sets, limiting real-world applicability. Some resort to synthetic or scene-specific data, further constraining generalization. In contrast, unsupervised or zero-reference approaches~\cite{guo2020zerodce,jiang2021enlightengan}  require no paired supervision, thereby broadening training data diversity and boosting robustness to unseen conditions. Recent flow-based methods have explored structure-aware and information-preserving strategies for low-light enhancement. UPT-Flow~\cite{xu2025UPT-Flow} introduces uncertainty priors and transformer-based normalization for improved robustness. JTE-CFlow~\cite{hu2024JTE-CFlow} targets missing pixels using joint attention and cross-map coupling. ZMAR-SNFlow~\cite{hu2024ZMARSNFlow} leverages semantic priors and zero-map guidance to enhance structure and color fidelity.

%Meanwhile, increasing efforts have been made toward lightweight network design to improve deployment efficiency. RUAS \cite{liu2021RUAS} employs Retinex-inspired unrolling and architecture search to reduce model size, while SCI \cite{ma2022SCI} adopts a self-calibrated module for fast inference, making both approaches noteworthy lightweight solutions. However, they still face performance bottlenecks and limited efficiency in extremely resource-constrained scenarios.

Despite the progress in LLIE, most state-of-the-art methods still suffer from high parameter and computation costs, which limit their applicability in resource-limited environments. The widespread use of multi-branch and attention-driven designs, although effective, further exacerbates inference latency and memory consumption. Additionally, heavy reliance on specialized training datasets limits generalization, causing over-enhancement, color distortion, or artifacts under unseen conditions. Most existing methods prioritize visual quality but neglect efficiency \cite{li2021low}, leaving a gap for truly practical solutions that achieve real-time performance on mobile platforms.

To address these issues, the motivation of this work is to push the limits of model compactness while maintaining effective enhancement performance, to meet the deployment constraints of embedded and mobile platforms. Our goal is to explore the extreme lightweight boundary of LLIE models by minimizing the number of parameters and operations without sacrificing visual quality. In other words, we aim to develop a practical LLIE solution that is visually effective and computationally efficient enough to run in real time on resource-limited devices—bridging the gap between academic research and real-world application.

With the rise of embedded systems and mobile devices, the demand for efficient LLIE has surged, yet most existing methods rely on large neural networks that impede real-time deployment. This paper revisits the trade-off between visual quality and runtime efficiency by proposing LiteIE, the lightest unsupervised approach that removes dependence on large-scale labeled datasets, thereby enhancing adaptability. LiteIE explores an extremely lightweight, structure-insensitive backbone that preserves stable representations across diverse scenes. A multi-stage progressive feature aggregation further refines multi-scale features while preserving details. On a Snapdragon 8 Gen 3-based device, LiteIE achieves real-time 4K (30 FPS) enhancement, demonstrating its suitability for resource-constrained platforms.

In summary, the main contributions are as follows:

\begin{itemize}
    \item[1)] Ultra-light backbone. We design a network-agnostic feature extractor that uses only two convolutions (58 weights in total) to produce a compact enhancement tensor.
	
    \item[2)] Parameter-free Iterative Restoration Module (IRM). IRM reuses previously extracted features to progressively recover fine details lost in earlier enhancement steps, without introducing any additional learnable parameters.
	
    \item[3)] Multi-Scale Color Consistency Loss. To ensure stable and natural enhancement, we design MSCol Loss, which preserves local color structure while maintaining global color balance.
	
\item[4)] Extensive validation. LiteIE generalizes well across diverse low-light datasets and runs efficiently on mobile SoCs, enabling fast inference on resource-limited devices.

\end{itemize}

\begin{figure*}[t]
    \centering
    \includegraphics[width=1.0\linewidth]{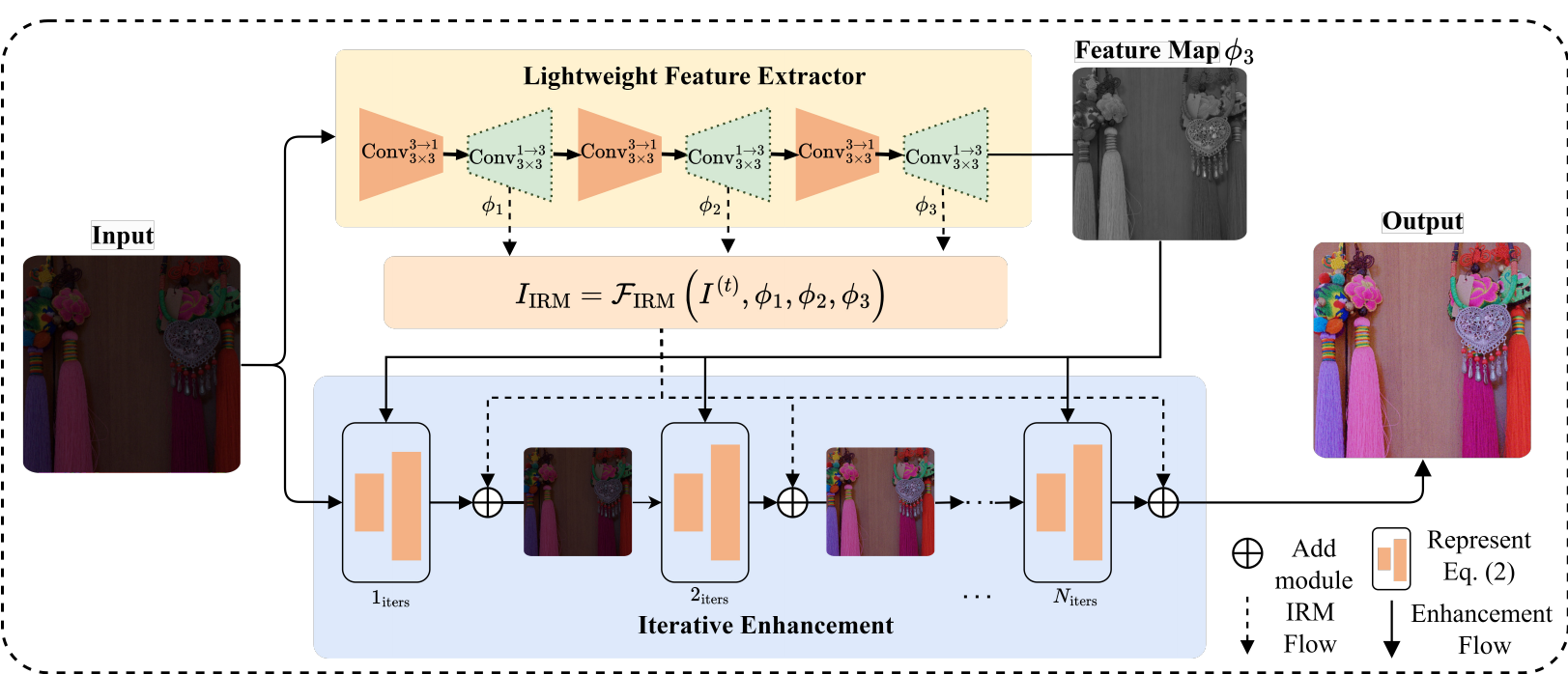}
    \caption{Architecture of the proposed LiteIE framework, consisting of a Lightweight Feature Extraction Network and an Iterative Restoration Module. The feature extractor uses two shared convolutional layers, repeated three times, with the final feature map as the enhancement matrix. Each iteration applies the Iterative Restoration Module to ensure detail recovery, improving overall image quality.}
    \label{fig:figure2}
    \vspace{-1.7em} % 缩短表格底部与正文之间的空白
\end{figure*}

\section{Related work}

Low-light image enhancement has emerged alongside the growing demand for improved visibility in challenging lighting conditions, undergoing continuous development over the years \cite{li2021low, guo2016lime, yang2020darkface}.

\textbf{\textit{A. Traditional Methods}}

Histogram equalization methods, such as adaptive histogram equalization  \cite{pizer1987adaptive} and contrast entropy-based approaches \cite{agaian2007transform}, are fundamental techniques for enhancing global contrast in low-light images by redistributing pixel intensities. Retinex theory \cite{1971retinex} forms the theoretical basis for many LLIE methods by decomposing an image into illumination and reflectance layers. Building on this, LR3M \cite{ren2020lr3m} introduces low-rank priors for robust decomposition, RRNet \cite{jia2023reflectance} emphasizes reflectance enhancement via re-weighting, and SDR \cite{hao2020low} proposes semi-decoupled modeling to better address spatially varying illumination.

\textbf{\textit{B. Learning-based LLIE Methods}}

\textbf{Retinex-based methods.} Retinex theory has been integrated into neural networks to mitigate traditional limitations, including noise amplification and color distortion. KIND++ \cite{zhang2021kind++} refines Retinex decomposition by jointly optimizing reflectance and illumination, enhancing visual quality without explicit supervision but with increased computational cost. URetinex-Net \cite{wu2022uretinex} utilizes a Retinex-inspired deep unfolding mechanism to enhance illumination and reflectance while managing computational complexity. RUAS \cite{liu2021RUAS} introduces a lightweight Retinex-inspired framework using unsupervised learning and attention mechanisms for adaptive enhancement with reduced computational requirements. SCI \cite{ma2022SCI} uses a self-calibrated illumination model to flexibly adjust illumination, mitigating over-enhancement and color distortion. Despite these advances, most Retinex-based deep learning methods still face challenges related to computational efficiency, which limits their real-time applicability.

\textbf{Curve-based Methods.} Curve-based methods treat low-light image enhancement (LLIE) as an adaptive curve estimation task, directly learning adjustment curves to correct brightness, contrast, and illumination. These approaches typically offer greater simplicity and computational efficiency compared to Retinex-based methods, making them suitable for real-time deployment on resource-limited devices. Zero-DCE~\cite{guo2020zerodce} first employed a lightweight CNN to estimate global illumination curves without paired supervision, enabling rapid inference, though its global curves struggle in unevenly illuminated scenes. Recent methods have explored more sophisticated curve formulations. ChebyLighter~\cite{pan2022chebylighter} introduces Chebyshev polynomials for multi-stage, localized curve refinement, providing enhanced local contrast control. Self-DACE~\cite{wen2023self} proposes a self-supervised approach combining adaptive curve estimation with a Retinex-inspired self-reference loss, significantly enhancing color fidelity without relying on labeled datasets. 

\textbf{Generative Methods.} Generative models like GANs and diffusion models are effectively applied in low-light image enhancement. GAN-based methods utilize an adversarial framework where a generator creates enhanced images and a discriminator distinguishes them from well-lit references. EnlightenGAN \cite{jiang2021enlightengan} targets brightness and contrast improvements, whereas CIGAN \cite{ni2022CIGAN} introduces interactive cycles to enhance robustness without paired supervision. Diffusion models have demonstrated promising performance in image generation, leading researchers to explore their use in low-light image enhancement.Diff-Retinex \cite{yi2023Diffretinex} integrates Retinex decomposition with diffusion processes for iterative latent feature refinement, while LightenDiffusion \cite{jiang2024lightendiffusion} aims at generating visually compelling results through denoising steps. Such advances highlight the versatility of generative models and suggest further potential for LLIE, albeit with attention to computational cost and training complexity.

% \textbf{Discussion.} Although learning-based approaches have achieved notable gains, heavy network architectures remain a bottleneck for real-time deployment. Moreover, reliance on specific training datasets can lead to artifacts in unseen environments. Our work tackles these issues with a minimalist yet effective framework, reducing complexity and dataset dependence to enable real-time LLIE on resource-constrained devices.

\begin{figure}
    
    \includegraphics[width=1.0\linewidth]{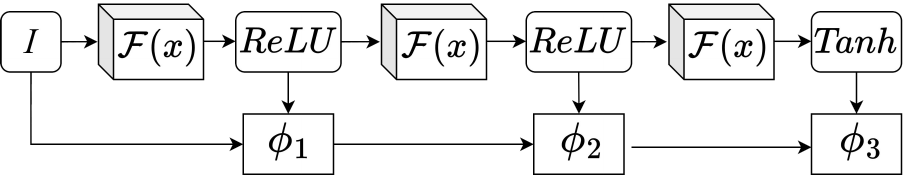}
    \caption{Two weight-sharing convolutional layers ($\text{Conv}_{3 \times 3}^{3 \rightarrow 1}$ and $\text{Conv}_{3 \times 3}^{1 \rightarrow 3}$) constitute the feature extraction operator $\mathcal{F}(x)$, progressively extracting image features $\phi_1$, $\phi_2$, and $\phi_{3}$ as references for restoration. The final feature map, $\phi_{3}$, serves as the output enhancement matrix.
    }
    \label{fig:features_extract}
    \vspace{-1.7em} % 缩短表格底部与正文之间的空白
\end{figure}

\section{Proposed method}
\label{sec:4}

We propose the Lightweight Iterative Enhancement and Restoration (LiteIE) framework for low-light image enhancement (see Fig.~\ref{fig:figure2}), consisting of two key components: a lightweight feature extraction module and an iterative restoration module. The feature extraction module efficiently captures multi-level features with minimal parameters, producing a final feature matrix that serves as the enhancement matrix. Meanwhile, the iterative restoration module iteratively refines image quality by leveraging the previously extracted features to ensure the preservation of fine details.

In the following subsections, we describe each component of the LiteIE framework in detail. Sec.~3.1 introduces the lightweight low-light enhancement module, Sec.~3.2 details the iterative restoration module that improves image quality while maintaining critical details, and Sec.~3.3 describes the unsupervised training loss functions used in the learning process.

%-------------------------------------------------------------------------
\subsection{Lightweight Low-Light Enhancement}

Inspired by curve-based enhancement methods \cite{guo2020zerodce,li2021zerodce++}, we reconceptualize low-light enhancement as the extraction of an enhancement matrix for images. The proposed Lightweight Iterative Enhancement and Restoration (LiteIE) network utilizes two convolutional layers to extract multi-level feature representations, employing the final feature matrix as the enhancement matrix to facilitate iterative enhancement and improve image quality. As shown in Fig.~\ref{fig:features_extract}, the network utilizes two convolutional layers, sized (\(\text{Conv}_{3 \times 3}^{3 \rightarrow 1}\) and \(\text{Conv}_{3 \times 3}^{1 \rightarrow 3}\)) for feature compression and reconstruction, repeated in a shared-weight manner across three stages to minimize computational complexity and parameter count. The feature extraction process can be described in three steps:

\begin{equation}
	\phi(I):\left\{
	\begin{array}{l}
		\phi_1=\operatorname{ReLU}(\mathcal{F}(I ; W)) \\
		\phi_2=\operatorname{ReLU}\left(\mathcal{F}\left(\phi_1 ; W\right)\right) \\
		\phi_{3}=\operatorname{Tanh}\left(\mathcal{F}\left(\phi_2 ; W\right)\right).
	\end{array}
	\right.
\end{equation}

\begin{figure}[!t]
	\centering
	
	% 图像部分
	\includegraphics[width=\linewidth]{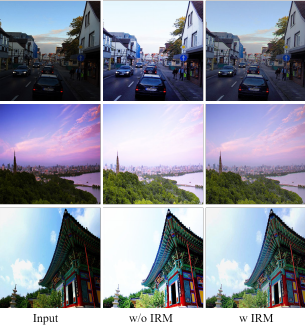}
	\caption{Visual comparison of models with and without the Iterative Restoration Module (IRM).}
	\label{fig:IRMablation}
	
	\vspace{0.8em}  % 控制图与表之间的距离
	
	% 表格标题与设置
	\captionsetup{type=table,
		justification=centering,
		labelsep=space,
		font=small}
	\caption{Quantitative performance comparison of models with and without the Iterative Restoration Module (IRM) across DICM~\cite{lee2013DICM}, LIME~\cite{guo2016lime}, and NPE~\cite{wang2013NPE} datasets.}
	\label{tab:comparison}
	
	% 表格内容
	\renewcommand{\arraystretch}{1.2}
	\resizebox{\linewidth}{!}{
		\begin{tabular}{lcccc}
			\toprule[1.2pt]
			\textbf{Model} & \textbf{PSNR$\uparrow$} & \textbf{SSIM$\uparrow$} & \textbf{MAE$\downarrow$} & \textbf{NIQE$\downarrow$} \\
			\midrule
			w/o IRM & 14.53 & 0.58 & 3893.78 & 3.98 \\ 
			w IRM & \textbf{19.04} & \textbf{0.61} & \textbf{1507.57} & \textbf{3.79} \\ 
			\bottomrule[1.2pt]
		\end{tabular}
	}
	
\end{figure}

The input image \(I\) is first processed by a shared-weight operator \(\mathcal{F}(\cdot\,;W)\), which consists of a convolution, batch normalization, and identity activation. This produces a set of non-linear feature maps \(\phi_1\), \(\phi_2\), and \(\phi_3\) that progressively capture multi-scale representations (see Fig.~\ref{fig:features_extract}).

In LiteIE, these feature maps guide the iterative enhancement process. In particular, \(\phi_3\) serves as a feature enhancement matrix that modulates the input \(I\), enabling spatially adaptive adjustments in low-light regions.

\begin{equation}
	\tilde{I}^{(t)} = I^{(t)} + \phi_3 \cdot \left( (I^{(t)})^2 - I^{(t)} \right)
\end{equation}
In this equation, \( I^{(t)} \) is the input at iteration \( t \), and \( \tilde{I}^{(t)} \) is the intermediate enhanced output. The term \( (I^{(t)})^2 - I^{(t)} \) acts as a non-linear contrast enhancer, boosting intensity in bright regions while suppressing dark ones. The modulation map \( \phi_3 \) provides spatial adaptivity, enabling stronger enhancement in underexposed areas while avoiding overexposure in well-lit regions.

The enhancement process is applied iteratively across multiple steps, typically over several iterations (e.g., 8). With each iteration, the input image undergoes progressive refinement, improving its brightness, contrast, and clarity. The spatial adaptiveness of  \(\phi_{3}\), derived from the shared convolutional layers, ensures that the enhancement is not uniform across the image but tailored to the local features. This iterative process avoids over-exposure or unnatural artifacts while ensuring the enhanced image maintains its natural textures and details. By the end of the iterative process, the final enhanced image exhibits significantly improved visibility and contrast, particularly in low-light conditions, while maintaining a natural, artifact-free appearance.
%\begin{figure*}[tb]
%	\centering
%	
%	\includegraphics[width=1.0\linewidth]{sec/Image/1.png}
%	\caption{Visual comparison of the LiteIE feature extraction network across channel configurations in Table~\ref{tab:Network_Channel_comparison}, demonstrating insensitivity to variations in convolutional architecture.}
%	\label{fig:figure5}
%\end{figure*}
%
%
%\begin{figure}[t]
%	\centering
%	\begin{minipage}[t]{1.0\linewidth}
    %		\includegraphics[width=\linewidth]{sec/Image/figure4.pdf}
    %		\caption{Visual comparison of models with and without the Iterative Restoration Module (IRM)}
    %		\label{fig:figure4}
    %	\end{minipage}%
%	\hfill
%	\begin{minipage}[t]{1.0\linewidth}
    %		\renewcommand{\arrayrulewidth}{1.0pt} % 设置线条宽度
    %		\small
    %		\resizebox{\linewidth}{!}{
        %			\begin{tabular}{c c c c c}
            %				\hline
            %				\textbf{Model} & \textbf{PSNR $\uparrow$} & \textbf{SSIM $\uparrow$} & \textbf{MAE $\downarrow$} & \textbf{NIQE $\downarrow$} \\ 
            %				\hline
            %				w/o IRM & 14.01 & 0.57 & 3893.78 & 4.07 \\ 
            %				w IRM & 19.04 & 0.61 & 1507.57 & 3.79 \\ 
            %				\hline
            %			\end{tabular}
        %		}
    %		\captionof{table}{Quantitative performance comparison of models with and without the Iterative Restoration Module (IRM) across DICM\cite{lee2013DICM}, LIME \cite{guo2016lime}, and NPE \cite{wang2013NPE} datasets.}
    %		\label{tab:comparison}
    %	\end{minipage}
%\end{figure}

%-------------------------------------------------------------------------
\subsection{Iterative Restoration Module}

While iterative enhancement effectively boosts brightness and contrast, excessive iterations may lead to the loss of fine details, such as the disappearance of clouds in the sky (see Fig.~\ref{fig:IRMablation}). To mitigate this, we introduce the Iterative Restoration Module. This module leverages three feature maps \(\phi_1\), \(\phi_2\), and \(\phi_{3}\) generated during enhancement, which capture rich semantic details. By reusing the feature matrices from the extraction phase, we restore details and refine the final output without adding new convolution layers or increasing parameters. The Iterative Restoration process is defined by:
%\begin{equation}
%    I_{\text{IRM}} = I + \sum_{i=1}^{3} \alpha_i \cdot \operatorname{Tanh}(\phi_i) \cdot \left( (I^{(t)})^2 - I^{(t)} \right) \cdot I_{\text{init}}.
%\end{equation}

\begin{equation}
	I^{(t+1)} = \tilde{I}^{(t)} + \sum_{i=1}^{3} \alpha_i \cdot \operatorname{Tanh}(\phi_i) \cdot \left( (\tilde{I}^{(t)})^2 - \tilde{I}^{(t)} \right) \cdot I_{\text{init}}
\end{equation}

In this process, \(I^{(t+1)}\) denotes the output of the IRM module at iteration \(t+1\), while \(\tilde{I}^{(t)}\) is the intermediate enhanced image from the previous step. The original low-light input is represented by \(I_{\text{init}}\). The feature maps \(\phi_1\), \(\phi_2\), and \(\phi_3\), extracted during the enhancement stage, provide semantic guidance for restoration. Scalar weights \(\alpha_1\), \(\alpha_2\), and \(\alpha_3\) control the contribution of each feature map in modulating the residual term, which adaptively restores local details and enhances contrast.

By utilizing the non-linear properties of Tanh, which bounds values between -1 and 1, the calibration ensures smooth and controlled adjustments, reducing the risk of over-enhancement. This approach is particularly useful for low-light image enhancement, as it maintains a delicate balance between improving brightness and contrast while preserving fine details and textures. The non-linear calibration ensures stability throughout the process, making the method both lightweight and highly effective for real-time applications on resource-constrained devices.

\begin{figure*}[t]
	\centering
	\includegraphics[width=0.99\linewidth]{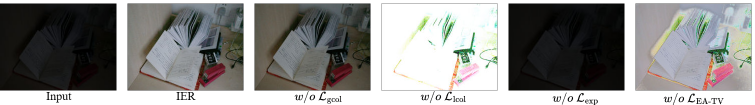}
	% \caption{Ablation study on loss functions. Removing the color consistency loss leads to strong color shifts (\textit{w/o Lgcol}, \textit{w/o lcol}), removing the exposure loss (\textit{w/o exp}) results in insufficient brightness, and discarding the smoothing loss (\textit{w/o smo}) introduces artifacts.}
	
	\caption{Ablation study on loss functions. Removing the color consistency loss (\textit{w/o} $\mathcal{L}{\text{lcol}}$,$\mathcal{L}{\text{gcol}}$) leads to strong color shifts , while excluding  the exposure loss (\textit{w/o} $\mathcal{L}{\text{exp}}$), images appear dim, and discarding the smoothing term (\textit{w/o} $\mathcal{L}_{\text{EA-TV}}$) introduces visible artifacts.}
	
	\label{fig:Ablation study}

\end{figure*}

\begin{figure}[t]
	\centering
	\includegraphics[width=0.99\linewidth]{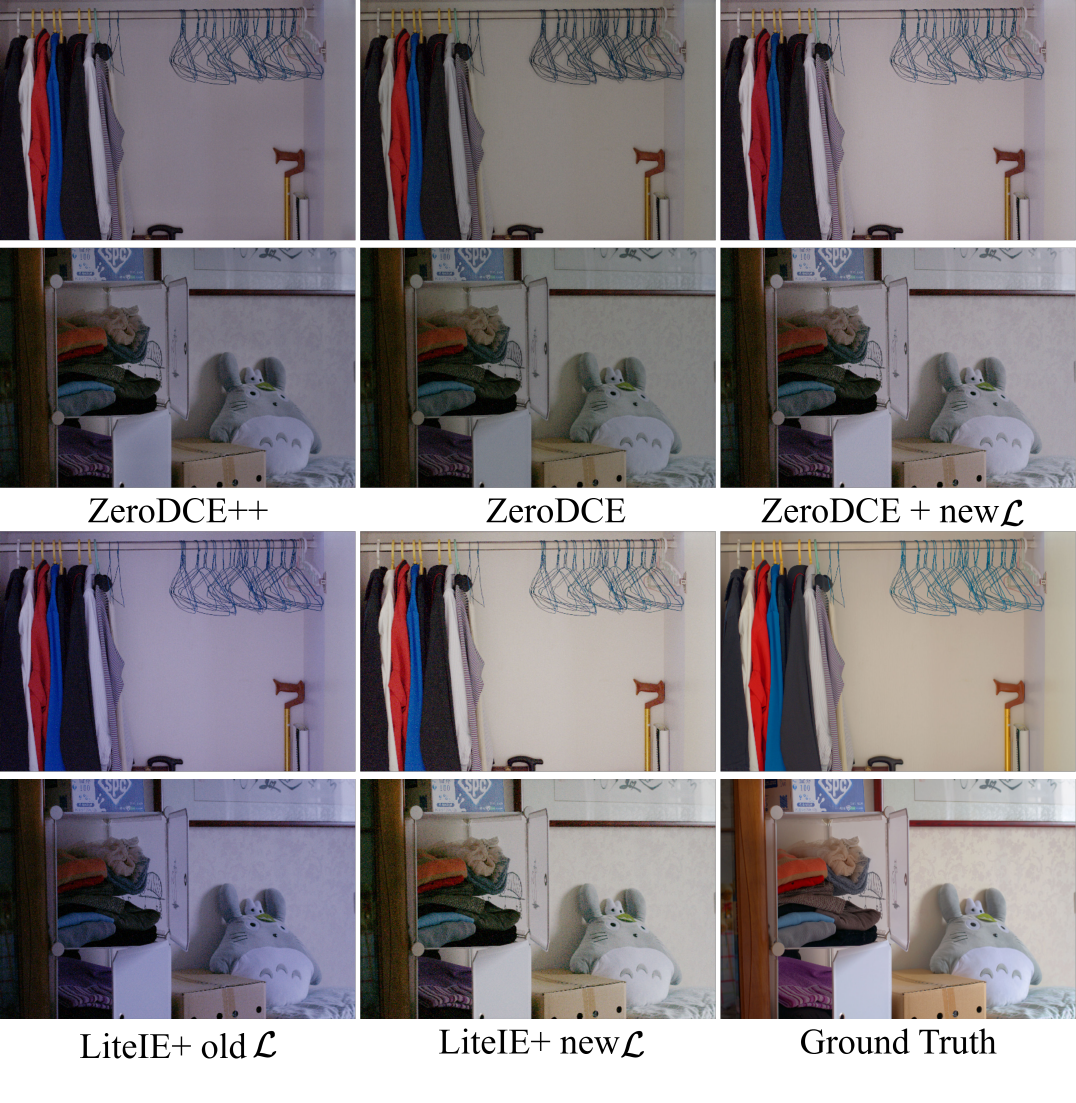}
	\caption{Visual comparison of representative methods. From left to right: Zero-DCE++, Zero-DCE, Zero-DCE + new loss, LiteIE + old loss, LiteIE + new loss, and ground truth. Zero-DCE++ shows noticeable color shifts. Incorporating the proposed global color loss improves color fidelity for both Zero-DCE and LiteIE. Our full model (LiteIE + new loss) produces the most natural and visually balanced results.}
	\label{fig:loss_Ablation}
\end{figure}

%\begin{figure*}[t]
%	\centering
%	% \begin{minipage}[t]{1.0\linewidth} % Adjust the width accordingly
    %		% 	\includegraphics[width=\linewidth]{sec/Image/figure9.png}
    %		% 	\caption{Precision-Recall (PR) curves with Average Precision (AP) values for different low-light image enhancement methods evaluated on the DarkFace\cite{yang2020darkface} dataset.}
    %		% 	\label{fig:PR}
    %		% \end{minipage}%
%	\hfill
%	\begin{minipage}[t]{1.0\linewidth} % Adjust the width accordingly
    %		\includegraphics[width=\linewidth]{sec/Image/figure10.pdf}
    %		\caption{Face detection results on the DarkFace dataset enhanced by different low-light methods.}
    %		\label{fig:DarkFace_results}
    %	\end{minipage}%
%	\hfill
%\end{figure*}

%-------------------------------------------------------------------------
\subsection{Unsupervised Training Loss Functions}
\textbf{Channel Adaptive Exposure Loss.} To ensure proper exposure while preserving color consistency, we propose the channel adaptive Exposure Loss (Exp Loss), which independently adjusts each color channel based on a dynamically computed reference from the original image, preventing unnatural shifts.
\begin{align}
    \mathcal{L}_{\mathrm{exp}} &= \sum\left(M_R^{\prime}-\alpha r_0 C_0\right)^2 
    + \sum\left(M_G^{\prime}-\alpha g_0 C_0\right)^2 \notag \\
    &\quad + \sum\left(M_B^{\prime}-\alpha b_0 C_0\right)^2,
\end{align}
where $M_R^{\prime}, M_G^{\prime}, M_B^{\prime}$ denote the mean RGB values of the enhanced image, and $r_0, g_0, b_0$ are the normalized RGB ratios of the original image. The chromatic consistency factor is defined as $C_0 = 1 - \sqrt{(r_0 - \frac{1}{3})^2 + (g_0 - \frac{1}{3})^2 + (b_0 - \frac{1}{3})^2}$, which measures deviation from a balanced color distribution, and $\alpha$ is a scaling parameter for exposure adjustment.

\textbf{Edge-Aware Total Variation Loss.} To address the issue of excessive smoothing in conventional TV loss \cite{li2021zerodce++}, we propose Edge-Aware Total Variation Loss (EA-TV Loss), which preserves edges by adaptively reducing smoothing in high-gradient regions.

\begin{equation}
    \mathcal{L}_{\text{EA-TV}} = \sum w_h \|\nabla_h x\|^2 + \sum w_w \|\nabla_w x\|^2,
\end{equation}
where $\nabla_h x$ and $\nabla_w x$ denote the horizontal and vertical gradients, respectively, and the edge-aware weights are defined as $
w = e^{-\beta |\nabla x|}
$
where $\beta$ controls the sensitivity to edges.

% \begin{figure*}[!t]
% 	\centering
% 	\includegraphics[width=0.99\textwidth]{sec/Image/figure5.pdf}
% 	\caption{Visual comparison of the LiteIE feature extraction network across channel configurations in Table~\ref{tab:Network_Channel_comparison}, demonstrating insensitivity to variations in convolutional architecture.}
% 	\label{fig:figure5}
% \end{figure*}

\begin{figure*}[t]
    \centering

    % 图片部分
    \includegraphics[width=0.99\textwidth]{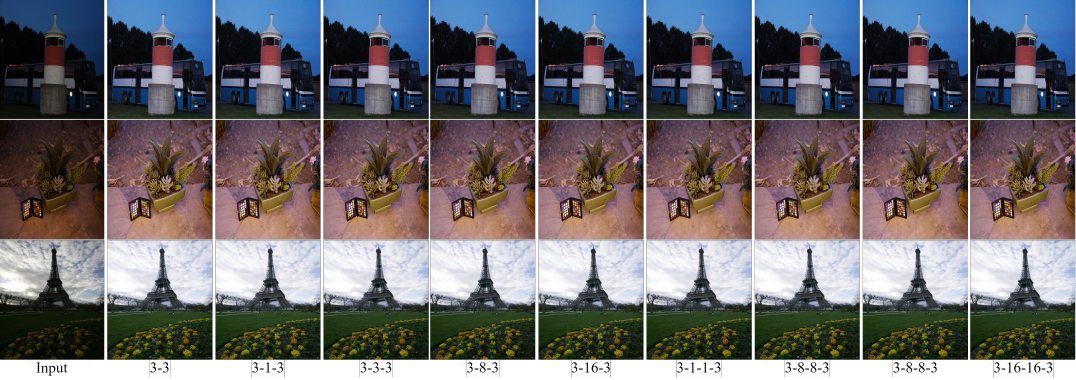}
    \caption{Visual comparison of the LiteIE feature extraction network across channel configurations in Table~\ref{tab:Network_Channel_comparison}, demonstrating insensitivity to variations in convolutional architecture.}
    \label{fig:figure5}

    \vspace{1em} % 可调节图和表之间的间距

    % 表格标题
    \captionsetup{type=table,
        justification=centering,
        labelsep=space,
        font=small}
    \caption{Comparative performance of LiteIE feature extraction network channel configurations for quality and efficiency. `-' indicates convolution layers, and numbers denote channel counts, e.g., '3-1-3' compresses channels from 3 to 1, then reconstructs to 3.}
    \label{tab:Network_Channel_comparison}

    % 表格内容
    \renewcommand{\arraystretch}{1.1}
    \resizebox{\textwidth}{!}{
    \begin{tabular}{cc|ccc|cccc}
        \toprule[1pt]
        \multicolumn{2}{c|}{\textbf{Setting for $\mathcal{F}(I, W)$}} 
        & \multicolumn{3}{c|}{\textbf{Quality}} 
        & \multicolumn{4}{c}{\textbf{Efficiency}} \\
        \cmidrule(r){1-2} \cmidrule(r){3-5} \cmidrule(r){6-9}
        \textbf{Blocks} & \textbf{Channels} 
        & \textbf{PSNR$\uparrow$} & \textbf{SSIM$\uparrow$} & \textbf{PI$\downarrow$} 
        & \textbf{Parameters} & \textbf{Model Size (KB)} & \textbf{FLOPs (G)} & \textbf{Runtime (s)} \\
        \midrule
        1 & 3-3 & 14.10 & 0.57 & 4.04 & 84 & 1.52 & 0.15 & 0.0008968 \\
        2 & 3-1-3 & 14.53 & 0.58 & 3.98 & 58 & 1.97 & 0.11 & 0.0009655 \\
        2 & 3-3-3 & 13.88 & 0.57 & 4.10 & 168 & 2.47 & 0.27 & 0.0010711 \\
        2 & 3-8-3 & 13.98 & 0.57 & 4.06 & 443 & 3.34 & 0.69 & 0.0012416 \\
        2 & 3-16-3 & 14.33 & 0.58 & 4.01 & 883 & 5.28 & 1.35 & 0.0012068 \\
        3 & 3-1-1-3 & 14.10 & 0.57 & 4.04 & 68 & 2.55 & 0.12 & 0.0011442 \\
        3 & 3-3-3-3 & 14.29 & 0.58 & 4.01 & 252 & 3.36 & 0.40 & 0.0013104 \\
        3 & 3-8-8-3 & 13.71 & 0.57 & 4.13 & 1027 & 6.30 & 1.57 & 0.0018968 \\
        3 & 3-16-16-3 & {14.62} & {0.58} & {3.96} & 3203 & 14.9 & 4.89 & 0.0016402 \\
        \bottomrule[1pt]
    \end{tabular}
    }
\end{figure*}

\begin{figure}[t]
	\centering
	\includegraphics[width=0.99\linewidth]{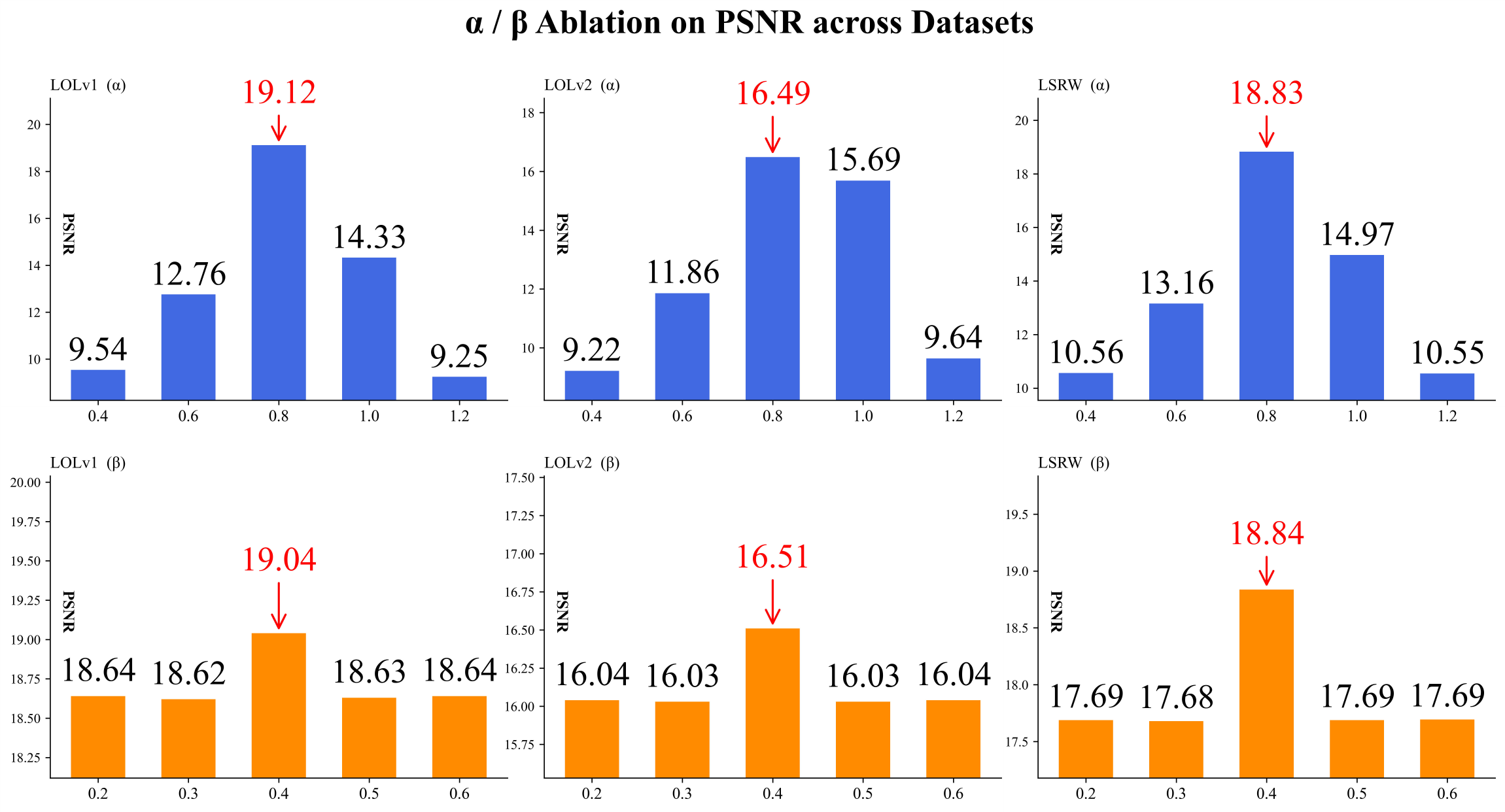}
	\caption{Ablation study of the loss weighting parameters \(\alpha/\beta\) on PSNR performance across LOL-V1~\cite{wei2018LOL}, LOL-V2, and LSRW~\cite{hai2023lsrw} datasets.}
	\label{fig:αaerfa_beita_Ablation}
\end{figure}

\textbf{Multi-Scale Color Consistency Loss.} To address the challenge of maintaining both local color consistency and global color balance, we propose MSCol Loss, which preserves local color structures while preventing global shifts, ensuring stable and natural enhancements.

\begin{align}
    \mathcal{L}_{\text{MSCol}} &=  \sum_{c \in \{r,g,b\}} (M_{\text{local}}^c - M_{\text{local, orig}}^c)^2  \notag \\
    &\quad +  \sum_{c \in \{r,g,b\}} (M_{\text{global}}^c - M_{\text{ref}}^c)^2,
\end{align}
\noindent where \( \mathcal{L}_{\text{MSCol}} \) represents the Multi-Scale Color Consistency Loss, with \( \lambda_{\text{local}} \) and \( \lambda_{\text{global}} \) controlling the balance between local and global constraints. \( M_{\text{local}}^c \) and \( M_{\text{local, orig}}^c \) denote the local mean color values of the enhanced and original image, ensuring regional color consistency, while \( M_{\text{global}}^c \) and \( M_{\text{ref}}^c \) enforce global color balance by aligning the overall distribution to a reference target.

\textbf{Total Loss.} The total loss is defined as:
\begin{equation}
    \mathcal{L} _{\text{total}} = \mathcal{L}_{\text{exp}} +  \mathcal{L}_{\text{EA-TV}} +  \mathcal{L}_{\text{MSCol}}.
\end{equation}
% where $\lambda_{\text{exp}}$, $\lambda_{\text{EA-TV}}$, $\lambda_{\text{MSCol}} $ are weighting factors controlling the contribution of each term.

% \begin{figure}[t]
%     \centering
%     \includegraphics[width=0.99\linewidth]{sec/Image/16.png}
%     \caption{Visual Comparison of Low-Light Image Enhancement Methods.}
%     \label{fig:figure8}

% \end{figure}

% \begin{figure}[t]
%     \centering
%     \includegraphics[width=0.99\linewidth]{sec/Image/161.png}
%     \caption{Visual Comparison of Low-Light Image Enhancement Methods.}
%     \label{fig:figure8}

% \end{figure}

\begin{figure*}[t]
    \centering
    \includegraphics[width=0.99\linewidth]{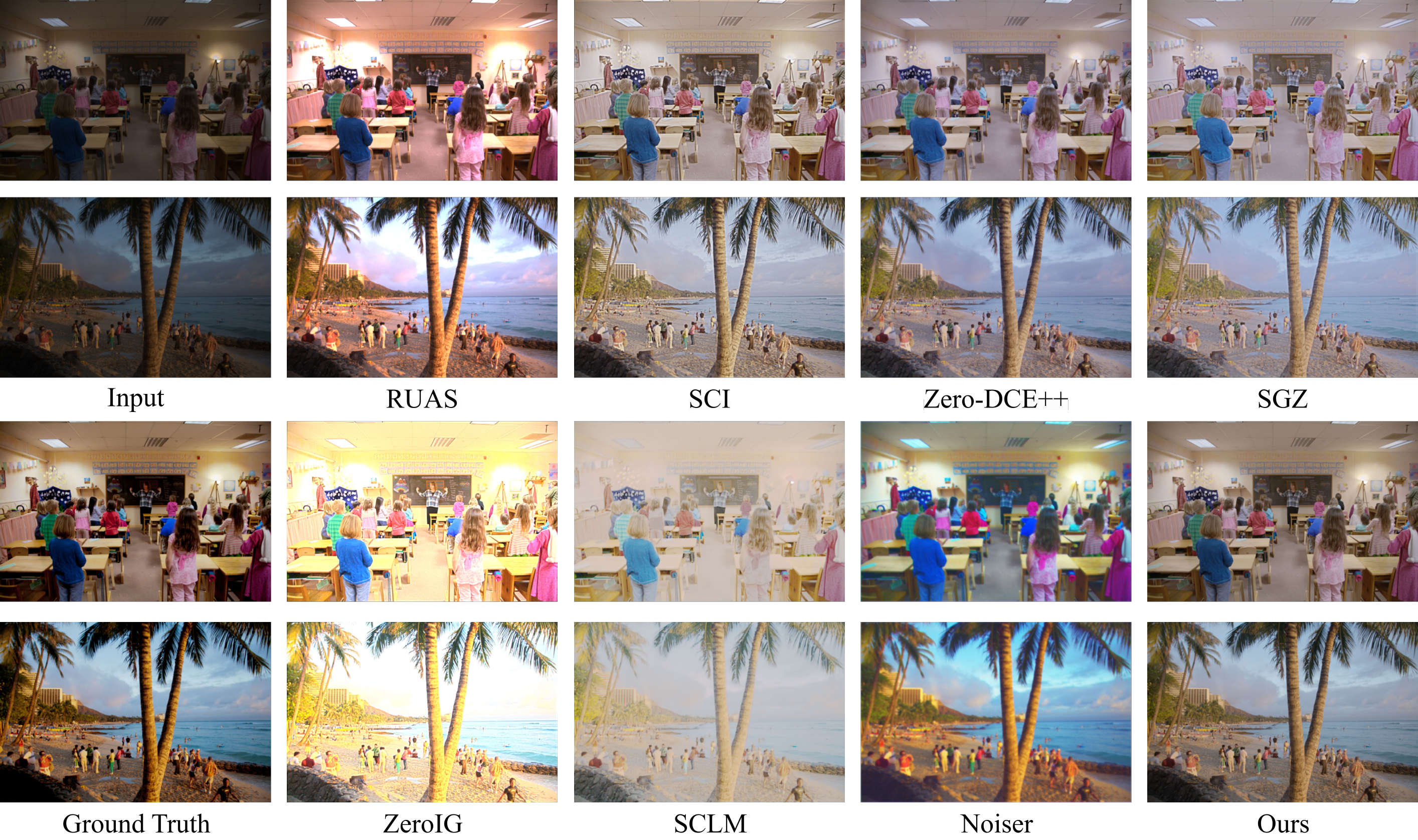}
    \caption{Visual Comparison of Low-Light Image Enhancement Methods.}
    \label{fig:figure8}

\end{figure*}

\begin{figure}[t]
    \centering
    \begin{minipage}[t]{1.0\linewidth}
        \includegraphics[width=\linewidth]{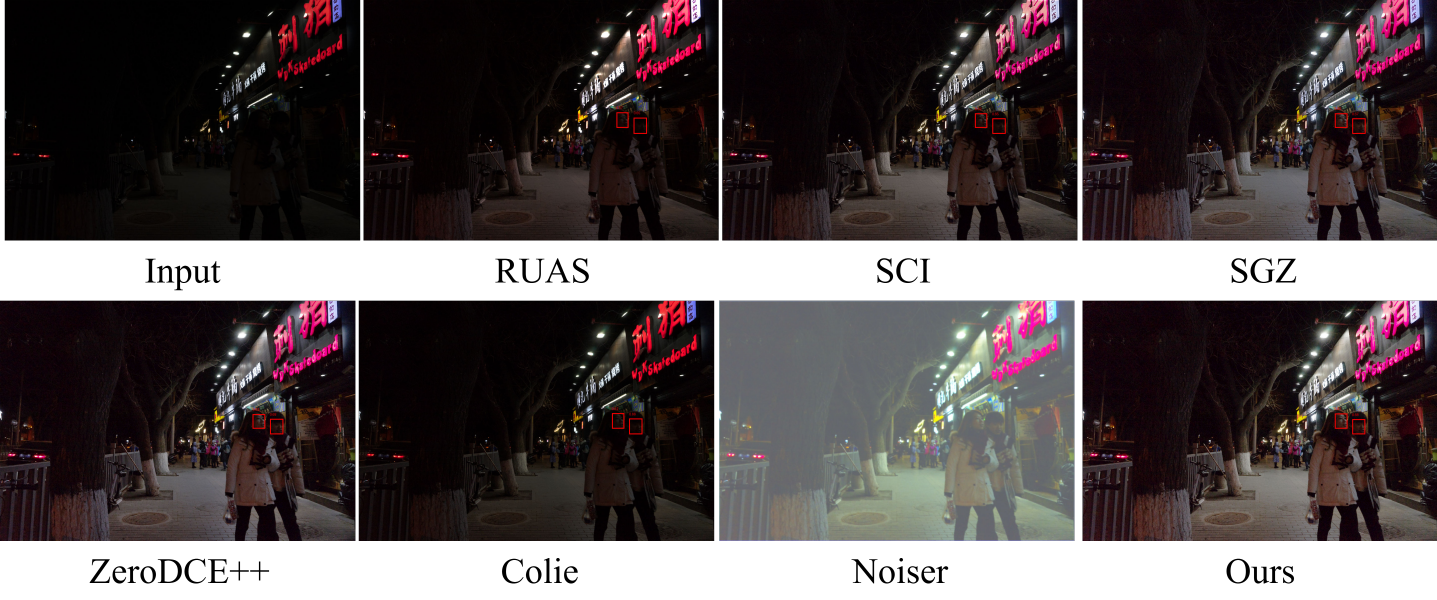}
        \caption{Detection performance visualization on the DarkFace dataset~\cite{yang2020darkface} with DSFD~\cite{li2019DSFD}.}
        \label{fig:mit-lsrw-visual}
    \end{minipage}

    \begin{minipage}{\columnwidth}
        \centering
        \scriptsize % 控制字体大小适应栏宽
        \captionof{table}{Performance comparison on MIT~\cite{fivekMIT}, LSRW-Huawei, and LSRW-Nikon~\cite{hai2023lsrw} datasets.}
        \label{tab:mit-lsrw}
        \resizebox{\linewidth}{!}{

        \begin{tabular}{l|cc|cc|cc}
            \toprule
            \multirow{2}{*}{\textbf{Methods}} & \multicolumn{2}{c|}{\textbf{MIT}} & \multicolumn{2}{c|}{\textbf{LSRW-Huawei}} & \multicolumn{2}{c}{\textbf{LSRW-Nikon}} \\
            \cmidrule{2-7}
            & \textbf{PSNR$\uparrow$} & \textbf{SSIM$\uparrow$} & \textbf{PSNR$\uparrow$} & \textbf{SSIM$\uparrow$} & \textbf{PSNR$\uparrow$} & \textbf{SSIM$\uparrow$} \\
            \midrule
            ZeroDCE~\cite{guo2020zerodce} & 15.97 & 0.76 & 16.34 & 0.46 & 15.03 & 0.41 \\
            ZeroDCE++~\cite{li2021zerodce++} & 16.31 & \textcolor{blue}{0.79} & 16.53 & 0.47 & \textcolor{blue}{15.61} & \textcolor{blue}{0.42} \\
            RUAS~\cite{liu2021RUAS} & 17.71 & 0.73 & 12.53 & 0.34 & 13.79 & 0.37 \\
            SCI~\cite{ma2022SCI} & 15.95 & 0.78 & 15.09 & 0.41 & 15.27 & 0.38 \\
            SGZ~\cite{zheng2022SGZ} & 14.50 & 0.74 & 17.04 & 0.47 & 15.73 & 0.41 \\
            CLIP-LIT~\cite{liang2023CLIPLIT} & \textcolor{blue}{17.32} & 0.80 & \textcolor{blue}{18.22} & 0.41 & 13.31 & 0.37 \\
            SCLM~\cite{zhang2023SCLM} & 9.99 & 0.50 & 14.03 & 0.41 & 14.05 & 0.41 \\
            ZeroIG~\cite{shi2024zeroIG} & 8.32 & 0.58 & 16.89 & 0.41 & 12.63 & 0.34 \\
            NoiSER~\cite{zhang2024NoiSER} & 16.91 & 0.72 & 15.79 & \textcolor{red}{0.52} & 15.58 & \textcolor{red}{0.44} \\
            \midrule
            \textbf{Ours} & \textbf{\textcolor{red}{17.73}} & \textbf{\textcolor{red}{0.81}} & \textbf{\textcolor{red}{18.54}} & \textbf{\textcolor{blue}{0.49}} & \textbf{\textcolor{red}{16.08}} & \textbf{\textcolor{red}{0.44}} \\
            \bottomrule
        \end{tabular}
        }
    \end{minipage}
    
\end{figure}

\begin{figure}[t]
    \centering

    % 图像部分
    \includegraphics[width=\linewidth]{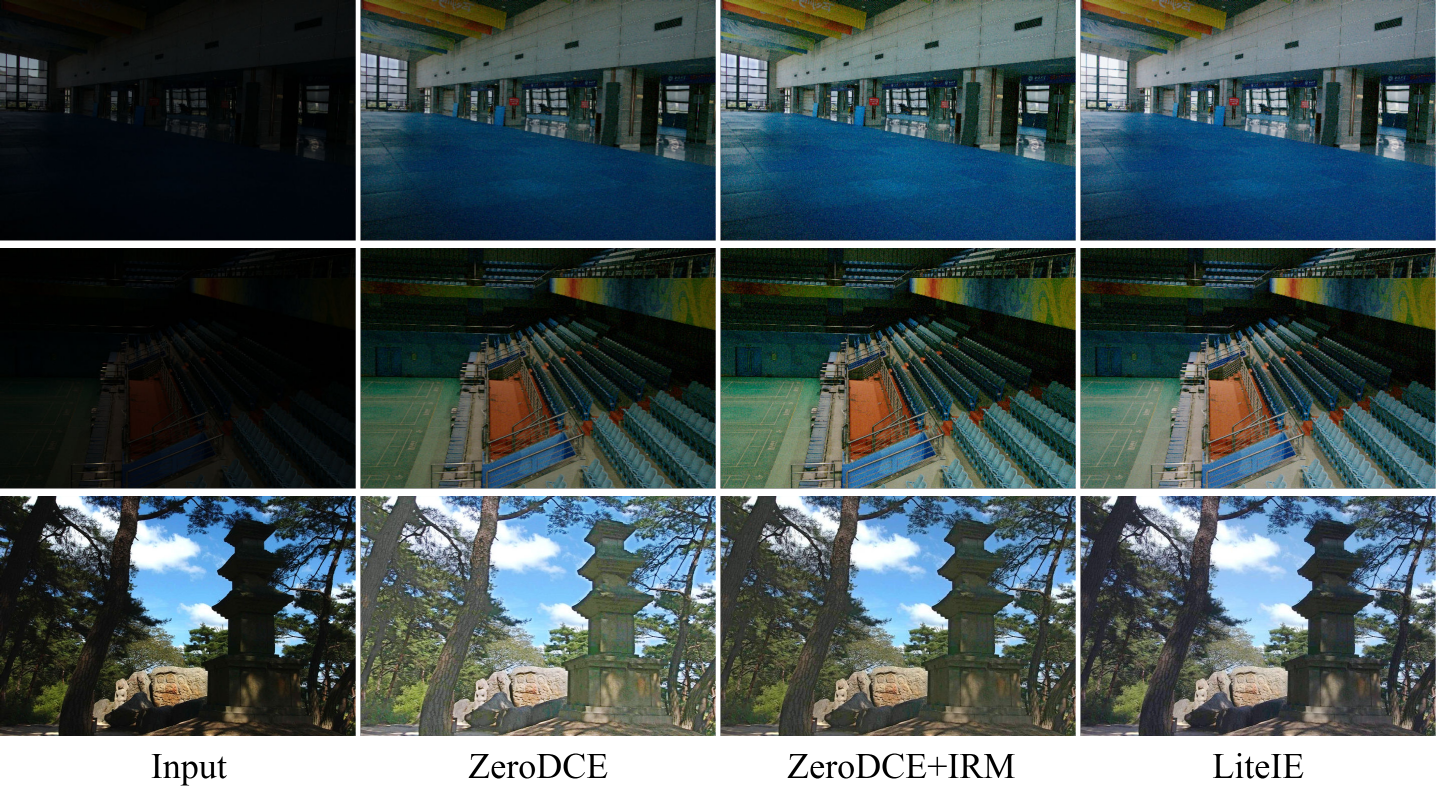}
    \caption{visual comparison of ZeroDCE with and without the iterative restoration module (IRM).}
    \label{fig:zero-dce+IRM}

    % 表格标题与设置
    \captionsetup{type=table,
        justification=centering,
        labelsep=space,
        font=small}
    \caption{Ablation study on backbone, loss function and IRM on LOL dataset \cite{wei2018LOL}.}
    \label{tab:ZeroDCE+IRM}

    % 表格内容
    \renewcommand{\arraystretch}{1.2}
        % \begin{tabular}{lccc}
        %     \toprule
        %     \textbf{Model} & \textbf{PSNR$\uparrow$} & \textbf{SSIM$\uparrow$} & \textbf{MSE$\downarrow$} \\
        %     \midrule
        %     ZeroDCE~\cite{guo2020zerodce} & 14.97 & 0.58 & 3231.67 \\
        %     ZeroDCE+IRM & 18.29 & 0.57 & 1605.62 \\
        %     LiteIE & \textbf{19.04} & \textbf{0.61} & \textbf{1507.57} \\
        %     \bottomrule
        % \end{tabular}
                \resizebox{\linewidth}{!}{

\begin{tabular}{lccccc}
\toprule
\textbf{Model} & \textbf{PSNR$\uparrow$} & \textbf{SSIM$\uparrow$} & \textbf{MSE$\downarrow$} & \textbf{\#Params$\downarrow$} & \textbf{FPS$\uparrow$} \\
\midrule
ZeroDCE\,\cite{guo2020zerodce}               & 14.97 & 0.58 & 3231.67 & 79K &  526\\
ZeroDCE + new $\mathcal{L}$                  & 17.46 & 0.59 & 1820.03 & 79K & 527 \\
ZeroDCE + IRM                                & 18.29 & 0.57 & 1605.62 & 79K &  442 \\
ZeroDCE + new $\mathcal{L}$ + IRM            & 18.92 & 0.60 & 1542.11 & 79K & 443 \\
\midrule
LiteIE + ZeroDCE $\mathcal{L}$               & 17.86 & 0.60 & 1673.07 & \textbf{58} & 1200 \\
\textbf{LiteIE (ours)}                       & \textbf{19.04} & \textbf{0.61} & \textbf{1507.57} & \textbf{58} & \textbf{1250} \\
\bottomrule
\end{tabular}
                }
\end{figure}

\begin{figure*}[t]
	\centering
	
	\begin{minipage}[t]{1.0\textwidth}
		\centering
		\includegraphics[width=0.99\textwidth]{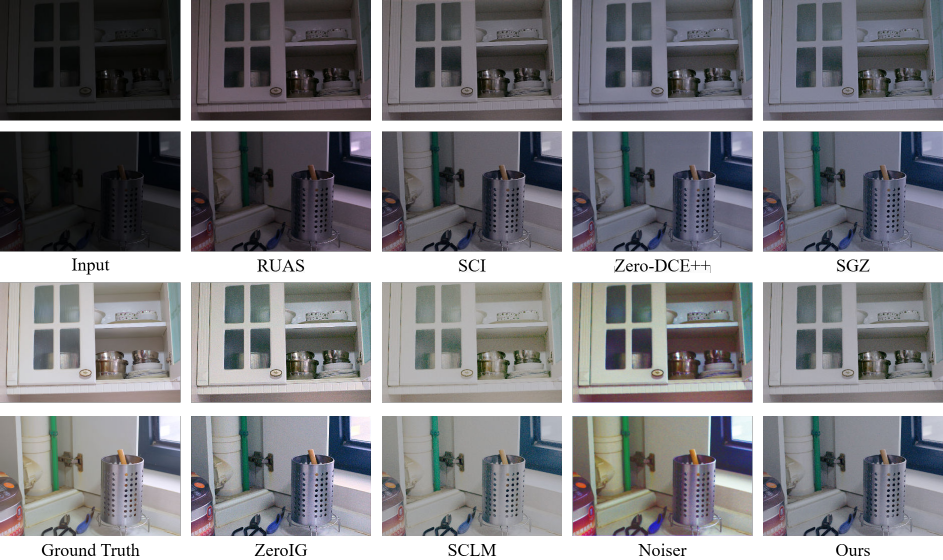}
		\caption{Visual comparison of low-light image enhancement methods. Results are compared across LOL \cite{wei2018LOL}, MEF \cite{ma2015perceptual}, NPE \cite{wang2013NPE}, and VV \cite{vv_dataset} datasets.}
		\label{fig:figure8}
	\end{minipage}

	\begin{minipage}[t]{\linewidth}
		%\centering
		\captionof{table}{Performance comparison of enhancement methods on the LOL-v1~\cite{wei2018LOL} dataset and efficiency metrics.}
		\label{tab:comparison_results}
		\renewcommand{\arraystretch}{1.1}
		\setlength{\tabcolsep}{2pt}
		\scriptsize % 控制字体
		\resizebox{\linewidth}{!}{%
			\begin{tabular}{l|c|c|c|c|c|c|c|c|c|c|c}
				\toprule
				\textbf{Methods} & \textbf{Venue} & \textbf{Params$\downarrow$} & \textbf{Model Size$\downarrow$} & \textbf{Latency(GPU)$\downarrow$} & \textbf{Latency(CPU)$\downarrow$} & \textbf{Latency(SoC)$\downarrow$} & \textbf{FPS(GPU)$\uparrow$} & \textbf{PSNR$\uparrow$} & \textbf{SSIM$\uparrow$} & \textbf{MAE$\downarrow$} & \textbf{LOE$\downarrow$} \\
				\midrule
				RRDNet~\cite{zhu2020RRDNet} & ICME’20 & 128.167 & - & $>$500 & $>$500 & $>$500 & $<$1 & 11.45 & 0.46  & 0.431 & 0.431 \\
				ZeroDCE~\cite{guo2020zerodce} & CVPR’20 & 79.416 & 30.63 & 2.539 & 1265.85 & 82.94 & 393  & 14.86 & 0.562 & 0.251 & 0.301 \\
				ZeroDCE++~\cite{li2021zerodce++} & TPAMI’21 & 10.561 & 312.51 & 1.974 & 12.15 & 57.91 & 506  & 14.71 & 0.46  & 0.243 & 0.359 \\
				RUAS~\cite{liu2021RUAS} & CVPR’21 & 3.438 & 30.63 & 4.421 & 96.52 & 109.15 & 226 & 16.40 & 0.503 & 0.202 & 0.312 \\
				SCI~\cite{ma2022SCI} & CVPR’22 & 0.258 & 42.93 & 4.773 & 57.43 & \textcolor{blue}{8.13} & 209 & 14.78 & 0.525 & 0.274 & \textcolor{blue}{0.297} \\
				SGZ~\cite{zheng2022SGZ} & WACV’22 & 10.561 & 51.07 & \textcolor{blue}{1.97}  & 9.29  & 22.11 & \textcolor{blue}{507} & 15.30 & 0.461 & 0.223 & 0.359 \\
				CLIP-LIT~\cite{liang2023CLIPLIT} & ICCV’23 & 278.79 & 1100.51 & 8.16  & 248.65 & 35.62 & 122  & 13.18 & 0.506 & 0.332 & 0.330 \\
				RDHCE~\cite{xia2023RDHCE} & IJCNN’23 & 296.22 & - & $>$500 & $>$500 & $>$500 & $<$1 & 17.11 & 0.482 & 0.191 & 0.308 \\
				SCLM~\cite{zhang2023SCLM} & TCSVT’23 & \textcolor{blue}{0.087} & 13.04 & 4.43  & \textcolor{blue}{8.86}  & -  & 225 & 10.68 & 0.424 & 0.495 & 0.312 \\
				COLIE~\cite{colie2024eccv} & ECCV’24 & 133.121 & - & $>$500 & $>$500 & $>$500 & $<$1 & 15.70 & 0.491 & 0.218 & 0.301 \\
				ZeroIG~\cite{shi2024zeroIG} & CVPR’24 & 86.572 & 349.62 & 6.82  & 700.1 & 210.06 & 146 & \textcolor{blue}{17.63} & 0.457 & 0.161 & 0.314 \\
				NoiSER~\cite{zhang2024NoiSER} & TPAMI’25 & 1.763 & \textcolor{blue}{8.45}  & 4.02  & 20.54 & - & 248 & 17.31 & \textcolor{red}{0.682} & \textcolor{blue}{0.143} & 0.299 \\
				\midrule
				\textbf{Ours} & - & \textbf{\textcolor{red}{0.058}} & \textbf{\textcolor{red}{2.69}} & \textbf{\textcolor{red}{0.97}} & \textbf{\textcolor{red}{6.69}} & \textbf{\textcolor{red}{3.55}} & \textbf{\textcolor{red}{1030}} & \textbf{\textcolor{red}{19.04}} & \textbf{\textcolor{blue}{0.607}} & \textbf{\textcolor{red}{0.141}} & \textbf{\textcolor{red}{0.29}} \\
				\bottomrule
			\end{tabular}%
		}
	\end{minipage}
	
\end{figure*}

\section{Evaluating Method Performance}
This section examines the lightweight LiteIE feature extraction and evaluates the IRM's impact on low-light image enhancement.

\subsection{Ablation Study on Loss Functions}

We conduct ablation experiments to evaluate the contribution of each loss component. As shown in Fig.~\ref{fig:loss_Ablation}, removing any single term leads to noticeable degradation. Excluding the global color loss results in severe color shifts, confirming its importance for maintaining global color balance. Removing the local color constraint causes tone inconsistency, especially in textured regions. Without the exposure loss, the output appears dim with insufficient brightness correction. Omitting the smoothness term introduces edge artifacts, highlighting its role in preserving structural integrity.
To further validate the proposed global color consistency loss, we perform visual and quantitative comparisons (Fig.~\ref{fig:loss_Ablation}, Table~\ref{tab:ZeroDCE+IRM}). Models trained with the original Zero-DCE loss—such as Zero-DCE++ and LiteIE—suffer from increasing color distortions (e.g., blue cast) as model capacity decreases. Adding our global loss significantly improves color fidelity. Our full model (LiteIE + new loss) achieves the most natural and balanced results, demonstrating the necessity of global color constraints, especially for lightweight architectures.

\begin{figure*}[t]
	\centering
	
	\begin{minipage}[t]{1.0\textwidth}
		\centering
		\includegraphics[width=0.99\textwidth]{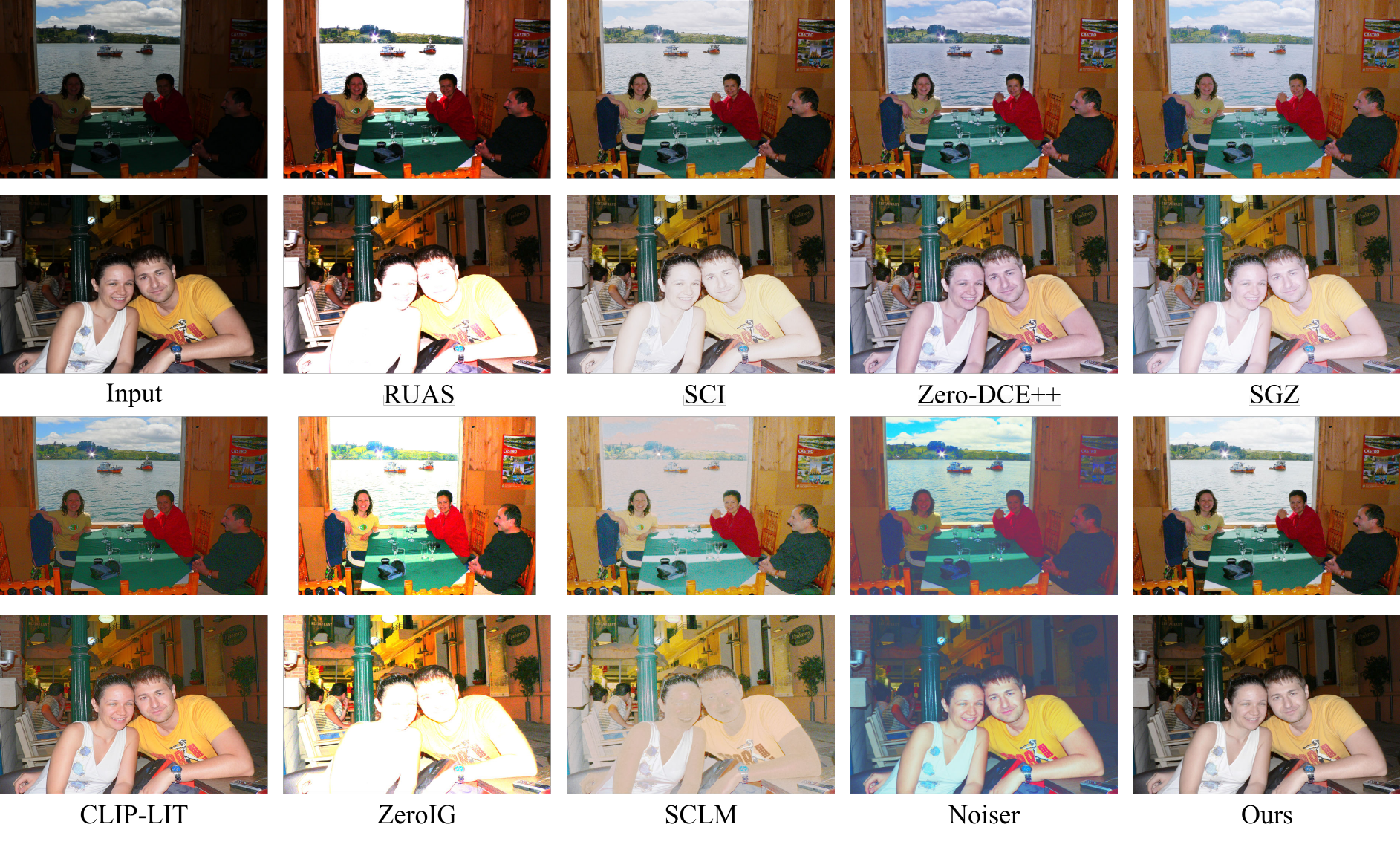}
		\caption{Visual comparison of low-light image enhancement methods across LOL \cite{wei2018LOL}, MEF \cite{ma2015perceptual}, NPE \cite{wang2013NPE}, and VV \cite{vv_dataset} datasets.}
		\label{fig:figure8}
	\end{minipage}

	\begin{minipage}[t]{\textwidth}
		\centering
		\captionsetup{type=table}  % 显式声明：这是表格，用于双栏兼容
		\caption{Performance comparison on four datasets using multiple metrics.(BRI: BRISQUE, CIQA: CLIP-IQA~\cite{wang2022exploring})}
		
		\renewcommand{\arraystretch}{1.2}
		\resizebox{\textwidth}{!}{%
			\begin{tabular}{c|cccc|cccc|cccc|cccc|c}
				\toprule
				\multirow{2}{*}{\textbf{Methods}} 
				& \multicolumn{4}{c|}{\textbf{LIME~\cite{guo2016lime}}} 
				& \multicolumn{4}{c|}{\textbf{NPE~\cite{wang2013NPE}}} 
				& \multicolumn{4}{c|}{\textbf{DARK FACE~\cite{yang2020darkface}}} 
				& \multicolumn{4}{c|}{\textbf{DICM~\cite{lee2013DICM}}} 
				& \textbf{Avg$\downarrow$} \\
				\cmidrule(lr){2-5} \cmidrule(lr){6-9} \cmidrule(lr){10-13} \cmidrule(lr){14-17}
				& \textbf{PI$\downarrow$} & \textbf{NIQE$\downarrow$} & \textbf{BRI.$\downarrow$} & \textbf{CIQA$\uparrow$}
				& \textbf{PI$\downarrow$} & \textbf{NIQE$\downarrow$} & \textbf{BRI.$\downarrow$} & \textbf{CIQA$\uparrow$}
				& \textbf{PI$\downarrow$} & \textbf{NIQE$\downarrow$} & \textbf{BRI.$\downarrow$} & \textbf{CIQA$\uparrow$}
				& \textbf{PI$\downarrow$} & \textbf{NIQE$\downarrow$} & \textbf{BRI.$\downarrow$} & \textbf{CIQA$\uparrow$}
				& \\
				\midrule
				% 原始数据保持不变
				ZeroDCE~\cite{guo2020zerodce} & \textcolor{blue}{2.79} & \textcolor{red}{3.99} & 18.48 & \textcolor{blue}{0.57} & \textcolor{blue}{2.86} & \textcolor{red}{3.95} & 15.64 & \textcolor{blue}{0.43} & 2.81 & 3.08 & 23.55 & 0.46 & \textcolor{red}{2.47} & 3.36 & 19.56 & 0.58 & 8.55 \\
				ZeroDCE++~\cite{li2021zerodce++} & 2.95 & 4.14 & 17.70 & \textcolor{red}{0.59} & 2.96 & 4.12 & \textcolor{red}{12.86} & \textcolor{blue}{0.43} & 2.86 & 2.87 & \textcolor{blue}{20.50} & \textcolor{blue}{0.49} & 2.71 & 3.48 & \textcolor{blue}{16.04} & 0.57 & \textcolor{blue}{7.76} \\
				RUAS~\cite{liu2021RUAS} & 3.15 & 4.46 & 24.20 & 0.56 & 3.87 & 5.66 & 41.17 & 0.32 & 3.17 & 3.59 & 18.78 & \textcolor{blue}{0.49} & 3.29 & 4.40 & 22.93 & 0.55 & 11.56 \\
				SCI~\cite{ma2022SCI} & 2.93 & 4.37 & \textcolor{blue}{17.43} & \textcolor{red}{0.59} & 2.94 & 4.14 & 16.49 & 0.42 & 3.17 & 3.56 & 21.18 & \textcolor{red}{0.51} & 2.89 & 3.84 & \textcolor{red}{14.68} & \textcolor{red}{0.62} & 8.14 \\
				SGZ~\cite{zheng2022SGZ} & 2.88 & 4.16 & 18.48 & 0.57 & 2.97 & 4.18 & 14.94 & 0.41 & 2.81 & 3.15 & 22.96 & \textcolor{blue}{0.49} & 2.57 & 3.60 & 19.59 & 0.57 & 8.52 \\
				CLIP-LIT~\cite{liang2023CLIPLIT} & 2.83 & 4.16 & 19.49 & 0.56 & 2.85 & 4.17 & 19.43 & 0.42 & 2.83 & 3.09 & 26.49 & 0.43 & 2.60 & 3.51 & 23.15 & 0.58 & 9.55 \\
				SCLM~\cite{zhang2023SCLM} & 3.01 & 4.03 & 17.58 & 0.45 & 3.21 & 4.32 & 15.06 & 0.34 & \textcolor{red}{2.20} & \textcolor{red}{2.76} & 24.69 & 0.32 & \textcolor{blue}{2.52} & \textcolor{blue}{3.29} & 21.32 & 0.41 & 8.67 \\
				ZeroIG~\cite{shi2024zeroIG} & 3.23 & 4.73 & 29.29 & 0.47 & 5.03 & 7.69 & 67.52 & 0.24 & 3.58 & 4.37 & 40.68 & 0.40 & 3.95 & 5.50 & 43.03 & 0.43 & 18.22 \\
				NoiSER~\cite{zhang2024NoiSER} & 4.96 & 5.39 & 37.11 & 0.39 & 5.01 & 5.34 & 34.18 & 0.31 & 4.19 & 4.01 & 26.80 & 0.32 & 4.38 & 4.37 & 31.81 & 0.27 & 13.96 \\
				
				\midrule
				\textbf{Ours} & \textbf{\textcolor{red}{2.65}} & \textbf{\textcolor{blue}{4.01}} & \textbf{\textcolor{red}{16.08}} & \textbf{\textcolor{red}{0.59}} & \textbf{\textcolor{red}{2.66}} & \textbf{\textcolor{blue}{4.08}} & \textbf{\textcolor{blue}{13.50}} & \textbf{\textcolor{red}{0.47}} & \textbf{\textcolor{blue}{2.65}} & \textbf{\textcolor{blue}{2.84}} & \textbf{\textcolor{red}{17.14}} & \textbf{\textcolor{blue}{0.49}} & \textbf{\textcolor{red}{2.47}} & \textbf{\textcolor{red}{3.02}} & \textbf{17.18} & \textbf{\textcolor{blue}{0.59}} & \textbf{\textcolor{red}{7.36}} \\
				
				\bottomrule
			\end{tabular}
		}
		\label{tab:comparison}
	\end{minipage}

\end{figure*}

\subsection{Hyperparameters Ablations}
To investigate the impact of the two hyperparameters introduced in the unsupervised loss formulation, we conduct ablation studies on $\alpha$ in the exposure loss $\mathcal{L}_{\text{exp}}$ and $\beta$ in the edge-aware total variation loss $\mathcal{L}_{\text{EA-TV}}$. These parameters control, respectively, the scaling of channel-adaptive exposure adjustment and the sensitivity to edges during smoothing.

\textbf{$\alpha$-Ablation (in $\mathcal{L}_{\text{exp}}$):} We test a range of fixed $\alpha$ values from 0.4 to 1.2 across three datasets: LOL-v1, LOL-v2, and LSRW. As shown in the top row of    Fig.~\ref{fig:αaerfa_beita_Ablation}, performance degrades significantly when $\alpha$ deviates from the optimal region, especially for extreme values (0.4 and 1.2). The best PSNR is consistently achieved at $\alpha = 0.8$, which balances brightness enhancement and color preservation without over-exposure or artifacts. Note that $\alpha$ is not a learnable parameter, but rather a fixed coefficient chosen through empirical evaluation.

\textbf{$\beta$-Ablation (in $\mathcal{L}_{\text{EA-TV}}$):} Similarly, we evaluate the effect of varying $\beta$ from 0.2 to 0.6. The results are visualized in the bottom row of Fig.~\ref{fig:αaerfa_beita_Ablation}. The PSNR scores are relatively stable across this range, with the best performance consistently observed around $\beta = 0.4$. Lower $\beta$ values lead to overly smoothed textures, while higher values reduce the ability to suppress noise in flat regions.

These results confirm that both $\alpha$ and $\beta$ play critical roles in guiding the training process. The selected values ($\alpha = 0.8$, $\beta = 0.4$) yield a good trade-off between exposure control, noise suppression, and edge preservation.

%-------------------------------------------------------------------------
\subsection{Extensibility of the Iterative Restoration Module}
Fig.~\ref{fig:IRMablation} compares models with and without IRM, illustrating how IRM effectively prevents color shifts and retains fine details (e.g., clouds). Meanwhile, Table~\ref{tab:comparison} provides quantitative metrics on three public datasets \cite{lee2013DICM,guo2016lime,wang2013NPE}, demonstrating that IRM significantly boosts PSNR and SSIM while reducing MAE and NIQE. These findings verify IRM’s ability to balance brightness enhancement and detail preservation across iterative processes.

Leveraging the same principle, we incorporate IRM into Zero-DCE \cite{guo2020zerodce} for comprehensive experiments.  Fig.~\ref{fig:zero-dce+IRM} shows that IRM mitigates haziness caused by repeated enhancements, improving clarity and color fidelity. Table~\ref{tab:ZeroDCE+IRM} further confirms performance gains in PSNR, SSIM, and MAE, underlining IRM’s versatility and impact on image quality. These results highlight the module’s scalability, indicating that IRM can be easily extended to other iterative enhancement algorithms seeking to preserve details in low-light scenes.

%-------------------------------------------------------------------------
\subsection{Insensitivity to Convolution Architecture}
In learning-based methods, adding more convolutional layers and channels often improves performance \cite{simonyan2014very}. However, the Lightweight Iterative Enhancement and Restoration (LiteIE) network reformulates low-light enhancement into a lightweight task of feature matrix extraction, significantly reducing the network size while retaining only minimal structures for feature extraction. By iteratively reusing the same feature extraction module $\mathcal{F}(I, W)$, multi-level feature representations are obtained, effectively reducing model parameters without compromising performance. We evaluated different configurations of feature extraction module $\mathcal{F}(I, W)$ (e.g., 3-3, 3-1-3, 3-16-16-3) as shown in Table~\ref{tab:Network_Channel_comparison}.

Fig.~\ref{fig:figure5} and Table~\ref{tab:Network_Channel_comparison} demonstrate that effective feature extraction can be achieved with only one or two convolutional layers. Metrics such as PSNR, SSIM, EME, and DE remain stable across configurations, confirming that minimal convolutions are sufficient for comparable enhancement quality. The 3-1-3 configuration, which consists of two convolutional layers (\(\text{Conv}_{3 \times 3}^{3 \rightarrow 1}\) and \(\text{Conv}_{3 \times 3}^{1 \rightarrow 3}\)), achieves quality comparable to more complex setups while using fewer parameters and having a smaller model size, ensuring computational efficiency. Additionally, Fig.~\ref{fig:figure5} visually confirms that the enhancement quality of 3-1-3 is similar to larger configurations, highlighting the effectiveness of this lightweight design.

\begin{table*}[t]
	\centering
	\caption{Runtime comparisons are conducted on Kirin 990 NPU and Snapdragon 8 Gen 3 NPU. We select four representative methods from Table~\ref{tab:comparison_results} for evaluation. The best results are bolded.}
	\resizebox{\textwidth}{!}{
		\begin{tabular}{l|cccc|cccc}
			\toprule
			\textbf{Method} 
			& \multicolumn{4}{c|}{\textbf{Kirin 990 5G SOC}} 
			& \multicolumn{4}{c}{\textbf{Snapdragon 8 Gen 3 }} \\
			& 1280$\times$720 & 1920$\times$1080 & 2560$\times$1440 & 3840$\times$2160 
			& 1280$\times$720 & 1920$\times$1080 & 2560$\times$1440 & 3840$\times$2160 \\
			\midrule
			ZeroIG~\cite{shi2024zeroIG}   & 585.4   & 1457.2 & 5484.7 & --      & 330.5  & 686.1  & 1190.3 & 2625.6 \\
			ZeroDCE~\cite{guo2020zerodce}& 445.8   & 954.3  & 1694.5 & --      & 196.2  & 434.9  & 728.4  & 1650.1 \\
			SGZ~\cite{zheng2022SGZ}      & 151.6   & 327.1  & 577.8  & 1341.4  & 70.5   & 150.2  & 261.7  & 577.9  \\
			SCI~\cite{ma2022SCI}         & 25.3    & 74.6   & 155.2  & 330.8   & 12.6   & 35.7   & 72.3   & 158.2  \\

			Ours    & \textbf{12.35} \textcolor{red}{$\uparrow$105\%} 
			& \textbf{26.9}  \textcolor{red}{$\uparrow$177\%} 
			& \textbf{43.9}  \textcolor{red}{$\uparrow$254\%} 
			& \textbf{87.0}  \textcolor{red}{$\uparrow$280\%} 
			& \textbf{5.87}  \textcolor{red}{$\uparrow$115\%} 
			& \textbf{11.2}  \textcolor{red}{$\uparrow$219\%} 
			& \textbf{20.2}  \textcolor{red}{$\uparrow$258\%} 
			& \textbf{37.4}  \textcolor{red}{$\uparrow$323\%} \\

			\bottomrule
		\end{tabular}
	}
	\label{tab:mobile_fps}
\end{table*}

\section{Experiments}

This section presents a comprehensive evaluation of the proposed LiteIE method. We first introduce the experimental setup, including datasets, evaluation metrics, and implementation details. We then compare the performance of LiteIE with several representative low-light enhancement algorithms through visual comparisons and quantitative metrics on both paired and unpaired datasets. 

To further assess the real-world applicability of LiteIE, we analyze its runtime performance across multiple hardware platforms, including desktop GPU, CPU, and mobile SoC. Finally, we conduct extensive qualitative and quantitative experiments to demonstrate the effectiveness and superiority of LiteIE across various low-light enhancement tasks.

\subsection{Experimental Settings}

We evaluate LiteIE on three paired-reference datasets (MIT~\cite{fivekMIT}, LSRW~\cite{hai2023lsrw}, and LOL~\cite{wei2018LOL}) and four unpaired datasets (LIME~\cite{guo2016lime}, NPE~\cite{wang2013NPE}, DARK FACE~\cite{yang2020darkface}, and DICM~\cite{lee2013DICM}). For quantitative assessment, we use full-reference metrics (PSNR, SSIM) for paired datasets, and no-reference metrics (NIQE, PI, BRI, and CIQA) for unpaired datasets. 

All image quality metrics are computed using the open-source IQA-PyTorch toolbox~\cite{pyiqa} to ensure consistency and reproducibility.
We also examine the runtime efficiency of LiteIE on three hardware platforms: a desktop GPU (NVIDIA RTX 4090), a CPU (Intel Xeon Silver 4310), and a mobile-end SoC (Snapdragon 8 Gen 3, tested on OnePlus 12). Additionally, we assess the practical utility of LiteIE by evaluating its effect on a downstream task, i.e., face detection on the DARK FACE dataset~\cite{yang2020darkface}, using the DSFD detector~\cite{li2019DSFD}.All training and inference experiments were conducted on a single NVIDIA RTX 4090 GPU with 24 GB memory.

%-------------------------------------------------------------------------
\subsection{Subjective Visual Tests}

We compared the proposed method with several state-of-the-art low-light enhancement techniques on the extremely low-light dataset LOL \cite{wei2018LOL} and complex lighting datasets DICM \cite{lee2013DICM} and LIME \cite{guo2016lime}. Fig.~\ref{fig:figure8} shows representative results, with each row depicting outputs from different algorithms. RUAS \cite{liu2021RUAS} shows minimal brightness improvement and often results in overexposure. SCI \cite{ma2022SCI} provides moderate improvements, but its results still suffer from poor consistency in challenging lighting conditions. Zero-DCE++ \cite{li2021zerodce++} demonstrates relatively consistent performance but lacks sufficient clarity in extremely low-light conditions. SGZ \cite{zheng2022SGZ} and CLIP-LIT \cite{liang2023CLIPLIT} both improve brightness but struggle with excessive color distortion. Compared to these methods, LiteIE not only enhances brightness and clarity in low-light conditions but also maintains accurate color under complex lighting. It provides superior light-shadow consistency and color accuracy, ensuring high visual quality across varied scenes.

%-------------------------------------------------------------------------
\subsection{Quantitative Tests}
We performed a thorough performance assessment of LiteIE on both paired datasets (MIT, LSRW-Huawei, LSRW-Nikon) and unpaired datasets (LIME, NPE, DARK FACE, DICM). As shown in Table~\ref{tab:mit-lsrw} , Table~\ref{tab:comparison_results}, and Table~\ref{tab:comparison}, LiteIE consistently achieves top performance in terms of PSNR and SSIM across all paired datasets, reflecting its strong capacity to restore both structural integrity and perceptual quality under low-light conditions.
In the unpaired setting, perceptual quality is evaluated using no-reference metrics, including PI, NIQE, BRI, and CIQA. LiteIE achieves the best average performance across all metrics, ranking first in PI and BRI, and among the top in NIQE and CIQA. These results demonstrate that LiteIE not only enhances visual clarity and contrast but also effectively preserves naturalness and suppresses artifacts. Overall, LiteIE consistently produces artifact-free, visually natural results across both controlled and complex lighting conditions.

%-------------------------------------------------------------------------
\subsection{Efficiency Analysis}

To assess runtime efficiency and deployment potential, we evaluated LiteIE on a server equipped with an NVIDIA RTX 4090 GPU and an Intel Xeon Silver 4310 CPU, as well as on smartphones powered by Kirin 990 5G and Snapdragon 8 Gen 3 SoCs. Table~\ref{tab:comparison_results} reports latency, FPS, model size, and parameter count. LiteIE achieves the lowest GPU latency of 0.97 ms, the highest throughput of 1030 FPS, and the smallest model size of 2.69 KB, outperforming all baselines. For mobile testing, all models were converted to TFLite format and accelerated with the GPU delegate. We tested four common resolutions, ranging from 720p to 4K, with results illustrated in Table~\ref{tab:mobile_fps}. LiteIE consistently delivers the highest FPS and is the only method that runs in real time at 4K (27.1 FPS on Snapdragon 8 Gen 3). In contrast, methods such as Zero-DCE\cite{guo2020zerodce} and Zero-IG~\cite{shi2024zeroIG} fail to handle 4K on-device due to heavy architectural designs. SGZ \cite{zheng2022SGZ} suffers from costly feature concatenation and resolution conversions, while SCI \cite{ma2022SCI} is constrained by iterative kernel-launch overhead. Benefiting from a minimal structure with only two convolutions and no redundant operations, LiteIE offers a strong balance between speed and accuracy, making it highly suitable for real-time mobile deployment.

%-------------------------------------------------------------------------

\begin{figure}[t]
	\centering
	\begin{subfigure}{0.32\linewidth}
		\centering
		\includegraphics[width=\linewidth]{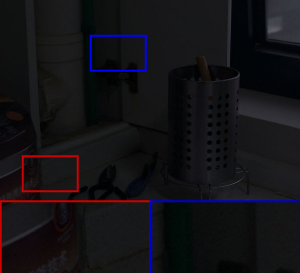}
		\vspace{-4mm}
		\caption*{(a) Input}
	\end{subfigure}
	\begin{subfigure}{0.32\linewidth}
		\centering
		\includegraphics[width=\linewidth]{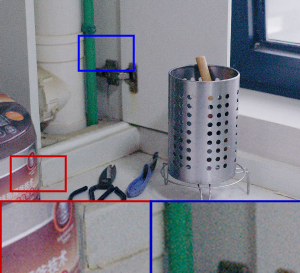}
		\vspace{-4mm}
		\caption*{(b) Ours}
	\end{subfigure}
	\begin{subfigure}{0.32\linewidth}
		\centering
		\includegraphics[width=\linewidth]{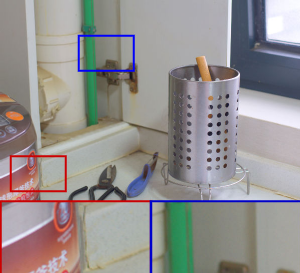}
		\vspace{-4mm}
		\caption*{(c) Ground Truth}
	\end{subfigure}
	
	\vspace{-2mm}
	\caption{
		Limitation in extremely low-light conditions: our method enhances structural visibility but also amplifies noise in dark areas (red boxes), with zoomed-in patches showing noise and slight color deviations.	}
	\label{fig:visual_comparison}
\end{figure}

\subsection{Limitations}
Although LiteIE demonstrates excellent performance across multiple datasets and mobile platforms, it still has certain limitations under challenging conditions. As shown in Fig.~\ref{fig:visual_comparison}, LiteIE effectively improves overall brightness and contrast while avoiding the noise amplification and color distortions seen in other methods (e.g., the middle result). However, it still struggles with fine detail recovery in some local regions. For example, in the blue-box area (e.g., the boundary between the green pipe and the wall), LiteIE produces slightly blurred textures, failing to fully restore the original structural details. In the red-box area (e.g., near the metallic container and label), while over-enhancement is successfully avoided, the result shows reduced sharpness in certain high-frequency regions compared to the reference.
In addition, although the iterative enhancement strategy prevents over-exposure and artifacts, it becomes less effective in extremely dark regions where the original signal is severely degraded. In such cases, the model may fail to reconstruct realistic colors and textures due to insufficient visual information. This limitation is particularly evident in indoor backlit scenes or night-time conditions with near-zero illumination.
%\titlespacing{\subsection}{0pt}{*1}{*1}
%
%
%
%%-------------------------------------------------------------------------
%\subsection{Dark Face Detection}
%
%
%
%
%
%In this section, we evaluated the performance of LiteIE and other competing algorithms on the DARK FACE \cite{yang2020darkface} dataset using DSFD\cite{li2019DSFD} as the base face detection model. DSFD, a state-of-the-art face detection approach, has demonstrated excellent performance on public benchmarks such as WIDER FACE \cite{yang2016wider}. For our evaluation, we randomly selected 1,600 images enhanced by each competing method from the DARK FACE dataset, totaling 12,800 images. We then computed precision-recall (P-R) curves and average precision (AP) scores at an IoU threshold of 0.5 for each method. As depicted in Fig.~\ref{fig:PR} and Fig.~\ref{fig:DarkFace_results}, the proposed LiteIE method achieves the highest AP score of 0.5061. URetinexNet shows overexposure, affecting detail retention and artifact control at high recall. In contrast, LiteIE (blue curve) maintains stable, high precision, demonstrating effective detail preservation and artifact suppression under complex low-light conditions.

\section{Conclusion}

In this work, we present LiteIE, a lightweight yet effective framework for low-light image enhancement. Motivated by the need to balance visual quality and computational efficiency, LiteIE adopts a minimalist design with only 58 parameters, leveraging two convolutional layers through recursive reuse. We systematically explored nine architectural variants and selected a 3-1-3 configuration, which offers a favorable trade-off between simplicity and performance. To address the degradation typically introduced by iterative enhancement, we propose an Iterative Restoration Module (IRM) that restores fine details by aggregating feature matrices across iterations without introducing additional network depth. Extensive experiments across multiple datasets and platforms demonstrate that LiteIE achieves state-of-the-art image quality while maintaining exceptional runtime efficiency and model compactness, making it highly suitable for deployment on edge and mobile devices.

\printcredits

\bibliographystyle{cas-model2-names}

\bibliography{cas-refs}


% Author biography without author photo.
% Author biography. Author biography. Author biography.
% Author biography. Author biography. Author biography.
% Author biography. Author biography. Author biography.
% Author biography. Author biography. Author biography.
% Author biography. Author biography. Author biography.
% Author biography. Author biography. Author biography.
% Author biography. Author biography. Author biography.
% Author biography. Author biography. Author biography.
% Author biography. Author biography. Author biography.

% \bio{cas-pic1}
% Author biography with author photo.
% Author biography. Author biography. Author biography.
% Author biography. Author biography. Author biography.
% Author biography. Author biography. Author biography.
% Author biography. Author biography. Author biography.
% Author biography. Author biography. Author biography.
% Author biography. Author biography. Author biography.
% Author biography. Author biography. Author biography.
% Author biography. Author biography. Author biography.
% Author biography. Author biography. Author biography.

% \bio{cas-pic1}
% Author biography with author photo.
% Author biography. Author biography. Author biography.
% Author biography. Author biography. Author biography.
% Author biography. Author biography. Author biography.
% Author biography. Author biography. Author biography.

\end{document}